\RequirePackage{fix-cm}

\documentclass{elsarticle}

\usepackage{xcolor}
\usepackage{hyperref}

\usepackage{graphicx}
\usepackage{amsmath,amssymb,bm} 
\usepackage{adjustbox} 
\usepackage{multirow} 

\begin{document}

\begin{frontmatter}

\title{Recognition Awareness: 
Adding Awareness to Pattern Recognition using Latent Cognizance}

\author[1]{Tatpong Katanyukul}
\ead{tatpong@kku.ac.th}

\author[2]{Pisit Nakjai\corref{mycorrespondingauthor}}
\cortext[mycorrespondingauthor]{Corresponding author}
\ead{mynameisbee@gmail.com}

\address[1]{Computer Engineering, Faculty of Engineering, Khon Kaen Univerity, Khon Kaen, 40002, Thailand}
\address[2]{Computer Science, Faculty of Science and Technology, Uttaradit Rajabhat University, Uttaradit, 53000, Thailand}

\begin{abstract}
	This study investigates an application of a new probabilistic interpretation of a softmax output 
to Open-Set Recognition (OSR).
	
	Softmax is a mechanism wildly used in classification and object recognition.
	However, a softmax mechanism forces a model to operate under a closed-set paradigm, i.e., to predict an object class out of a set of pre-defined labels.
	This characteristic contributes to efficacy in classification, but poses a risk of non-sense prediction in object recognition.
	Object recognition is often operated under a dynamic and diverse condition.
	A foreign object---an object of any unprepared class---can be encountered at any time. 
	OSR is intended to address an issue of identifying a foreign object in object recognition.
		
	Softmax inference has been re-interpreted with the emphasis of conditioning on the context.
	This re-interpretation and Bayes theorem have led to an approach to OSR, called Latent Cognizance (LC).
	LC utilizes what a classifier has learned and provides a simple and fast computation for foreign identification.  
	
	Our investigation on LC employs various scenarios,
	using Imagenet 2012 dataset as well as foreign and fooling images.
Its potential application to adversarial-image detection is also explored.
Our findings support LC hypothesis and show its effectiveness on OSR.
\end{abstract}


\end{frontmatter}

\section{Introduction}
\label{sec:introduction}

A well-adopted softmax function along with its accompanying cross-entropy loss has been introduced by Bridle\cite{Bridle1990} in 1990.
Since then, softmax has been proved effective and used extensively in classification.
%
With a proper setting, 
a softmax output converges to a class probability conditioned on the input.

Nevertheless, 
softmax limitations become more noticeable in object recognition \cite{ScheirerEtAl2013a, OpenMax2016},
where a chance to encounter an un-prepared class is common. 
Many approaches address this issue through open-set recognition (OSR).

While there are various approaches to OSR,
one potential is striking 
as it has been derived directly from a probabilitistic 
re-interpretation of a softmax output.
It is Latent Cognizance (LC) proposed by Nakjai and Katanyukul\cite{NakjaiKatanyukul2018a}.
LC was not originally intended for OSR.
It was to address the issue of un-prepared classes in hand-sign recognition,
but its underlying hypothesis is general.
Its mechanism can fit to various domains.
The new interpretation underlying LC has been verified
using synthetically traceable examples \cite{NakjaiEtAl2019b}.
LC applications have been shown to be effective in
hand-sign recognition\cite{NakjaiKatanyukul2018a}
and facial expression recognition\cite{AtsawaraungsukEtAl2021a}.

However, the recently introduced LC has not been adequately investigated for other domains.
An investigation of LC application to a more general domain, such as OSR, will allow a better insight into LC potential and the hypothesis behind it.
Our study here is set out to investigate an application of LC to OSR
and its related issues, such as an OSR evaluation metric,
a role of its base classifier,
and its potential for adversarial-image detection.

\section{Background}
\label{sec:background}

Despite that a softmax output
is conventionally viewed
as class probability, 
%
many literature\cite{GalGhahramani2016, OberdiekEtAl2018, Neal2018ECCV} have commented on a softmax output that often found uncorrelated to class probability,
especially when an input is ``foreign''. 
For conciseness, a foreign%
\footnote{
Previous studies have referred to 
foreign as either ``unknown'' or ``unseen''.}
%
 image will be referred  to an image of any class that has not been included in model preparation.

Based on this observation,
a new interpretation (\textsection\ref{sec:newinterpretation}) of a softmax output
is proposed \cite{NakjaiKatanyukul2018a}.
The new interpretation reveals 
a relation between penultimate values and posterior probabilities,
which in turn
has led to an invention of Latent Cognizance (LC, \textsection\ref{sec:LC}).
LC exploits what already learned in a deep network
to estimate a probability if an image is ``domestic''---i.e., of any category used in model preparation.
This allows its application to open-set recognition (\textsection\ref{sec:OSR}).



\subsection{Probabilistic Interpretation of Softmax Output}
\label{sec:newinterpretation}

A softmax function is commonly employed for 
multi-class classification.
Softmax computation \eqref{eq: softmax} is performed at the last calculation of a classifier.
To classify an input $\mathbf{x}$ into $1$ of $K$ classes, 
it is to compute the predicted class output $y = \arg\max_{k \in \{1, \ldots, K\}} y_k$.
Denote 
a softmax output $\mathbf{y} = [y_1, \ldots, y_K]^T$,
where
%
\begin{align}
y_k = \mathrm{softmax}_k(\mathbf{a})= \frac{\exp(a_k)}{\sum_{i=1}^K \exp(a_i)}
,
\; \text{for} \; k = 1, \ldots, K
\label{eq: softmax}.
\end{align}

A logit or penultimate vector $\mathbf{a} = [a_1, \ldots, a_K]^T = f'(\mathbf{x}, \mathbf{w})$,
when 
$\mathbf{w}$ represents network parameters
and
$f'$ is a network computation prior to the softmax.
The realization of $f'$ depends on a specific network configuration,
while values of $\mathbf{w}$ are obtained through a training process.

Softmax 
regulates 
$y_k \in [0,1]$ 
and $\sum_{k=1}^K y_k = 1$ for all $k$'s.
%
Bridle\cite{Bridle1990} shows that 
a well-trained classifier has its
softmax output converged to the posterior class probability:
$y_k \equiv p(y = k|\mathbf{x})$.
%
%
%
Softmax is effective 
in classification.

Extended beyond classification, 
softmax output 
is often found unrelated to class probabilities
when an input is foreign%
\cite{GalGhahramani2016, OberdiekEtAl2018, Neal2018ECCV}%
.
%
This observation contradicts the carried-over perception:  
$y_k \equiv p(y = k|\mathbf{x})$.
Thus, 
the softmax output is reinterpreted 
as a class probability conditioned on a given domestic input \cite{NakjaiKatanyukul2018a}:
\begin{align}
y_k \equiv p(y=k|\mathbf{x}, s)
\label{eq: LC interpretation},
\end{align}
where $s$ indicates the context 
that $\mathbf{x}$ is domestic or $s \equiv (y=1 | y=2 | y=3 | \ldots | y=K)$.
This interpretation emphasizes the domestic condition $s$.

The realization of this conditioning
is not unique to LC. 
Various literature\cite{YOLO2016, Yoshihashi_2019_CVPR} have stated this conditioning.
This interpretation has also been verified 
using a set of traceable (but small) examples \cite{NakjaiEtAl2019b}.

\subsection{Latent Cognizance}
\label{sec:LC}

Based on \eqref{eq: LC interpretation} and Bayes' theorem, the softmax output can be written as:
\begin{align}
y_k \equiv p(y=k|\mathbf{x},s) = \frac{p(y=k,s|\mathbf{x})}{\sum_{i=1}^K p(y=i,s|\mathbf{x})} 
\label{eq: LC bayes}.
\end{align}

Conferring \eqref{eq: softmax} to \eqref{eq: LC bayes},
the relation \eqref{eq: LC relation} is found:
\begin{align}
\frac{\exp(a_k)}{\sum_i \exp(a_i)}
= \frac{p(y=k,s|\mathbf{x})}{\sum_i p(y=i,s|\mathbf{x})}
\label{eq: LC relation}.
\end{align}
Consider similar patterns on both sides of \eqref{eq: LC relation},
a relation between penultimate values 
and the probabilities
is hypothesized:
%
given a well-trained network,
the penultimate vector 
relates to posterior probability 
through function $\tilde{h}_k(\mathbf{a}) = p(y=k,s|\mathbf{x})$.
To lessen a burden on enforcing probabilistic properties,
it is more convenient to work with a function whose value just correlates to the probability
rather than working directly with $\tilde{h}_k(\mathbf{a})$.
%
Assume that there exists
a monotonic function $g(\cdot)$ such that 
$g(a_k) \propto p(y=k,s|\mathbf{x})$.
Thus, marginalization reveals
%
\begin{align}
\sum_{k=1}^{K} g(a_k) \propto p(s|\mathbf{x})
\label{eq: LC sum g}.
\end{align}

Function $g(\cdot)$ is called a cognizance function.
A marginalized cognizance $\sum_{k=1}^{K} g(a_k)$ quantifies the degree of $\mathbf{x}$ being domestic: the lower value is, the more likely $\mathbf{x}$ is foreign.

\subsection{Open-Set Recognition}
\label{sec:OSR}

Open-Set Recognition (OSR) can be viewed as a mapping task, $F^{OSR}: \mathbf{x} \rightarrow \{0, 1, \ldots, K\}$,
where 
an input image $\mathbf{x}$
is mapped to a label index.
Indices $1, \ldots, K$ are associated to $K$ pre-defined classes used in model preparation/training process.
These prepared classes are called domestic.
Label index $0$ represents any class not used in the training process,  collectively called a foreign class.
In practice, a foreign class can represent multiple classes.

OSR is different from 
object detection \cite{ViolaJones2001, R-CNN2014, YOLO2016}.
As Scheirer et al\cite{ScheirerEtAl2013a} have pointed out,
an object-detection model is usually trained on images containing both positive and negative examples,
while OSR is to identify if an image is foreign or domestic
without negative examples.
If an image is domestic, OSR also has to recognize its category.

With this outline, OSR can be viewed as 
a conventional object recognition with additional novelty detection capability.
Many novelty detection methods \cite{PimentelEtAl2014} rely on some kinds of a distance-based scheme, using a distance between the input and the  training samples as a cue.
This distance-based approach
is often characterized by a search over training examples (or their representatives).
This search is computationally expensive.
%
%
%
In additions, designed solely for novelty detection, 
these methods
do not scale well to OSR.
Neither does it
exploit 
a well-trained classifier,
which is available in OSR settings.

To address the issue from OSR perspective, 
Scheirer et al\cite{ScheirerEtAl2013a} formalize
OSR definition and propose 
a concept of an open-space risk 
as well as
a SVM-based one-vs-set machine.
Instead of using a single hyperplane 
as in 
one-class SVM 
\cite{ScholkopfEtAl1999},
the one-vs-set machine
uses two hyperplanes to bound the domestic samples,
in order to minimize the open-space risk.
Later approaches\cite{OpenMax2016, RuddEtAl2018} rely more on statistic models
and are reported with better foreign identification.

OpenMax\cite{OpenMax2016} employs statistic models 
to identify foreign samples: a sample whose 
all penultimate values are too different from their corresponding statistics is likely to be foreign.
Specifically, OpenMax operates in two phases:
(phase 1) meta-recognition calibration
is to learn domestic statistics
and
(phase 2) class probability re-adjustment
is to re-adjust softmax output and estimate a probability of being foreign, based on statistics learned in phase 1.
In addition, OpenMax employs thresholding to overrule
domestic prediction if its probability is below the threshold.
Weibull distribution is used to model each intra-class distance.
Similar to OpenMax, Extreme Value Machine (EVM\cite{RuddEtAl2018}) also uses statistic models.
But rather than relying on a base-classifier like what OpenMax does,
EVM proposes to use each class statistic model for both classification and foreign identification.
The result allows a training process to be done at once---it learns both classification and foreign identification in one training process.
In addition, EVM can easily add 
a statistic model for each new class found.
However, this comes with a cost of using a less efficient classification inference. 
EVM has to search over all non-redundant training examples to complete the task.
Note that both OpenMax and EVM implicitly imply a uni-modal distribution of intra-class distances.

Resorting to a generative model, 
Neal et al\cite{Neal2018ECCV} 
have used counterfactual images as foreign samples
 to train a new classifier accounting for $K + 1$ classes: $K$ domestic classes and one additional class representing any foreign class.
The counterfactual images are supposed to look closest to the domestic images, but not belong to any domestic class.
With conjecture that 
counterfactual images lie just outside the ideal decision boundaries,
using them in training could help tighten decision boundaries of the new classifier.

Pivotal to their approach, it is how to create counterfactual images.
Neal et al
have prepared the encoder and the generator,
then used them in the synthesis of counterfactual images.
The encoder and the generator
are obtained through a generative adversarial framework  with reconstruction loss \cite{GulrajaniNIPS2017, SalimansNIPS2016, BerthelotSM17, ZhuEtAl2017iccv}.
That is, given a set of training images $\mathbf{X}$, discriminator $D$ and generator $G$ (along with encoder $E$) are trained in alternating steps with discriminator and generator losses as
$L_D = \sum_{\mathbf{x} \in \mathbf{X}} D(G(E(\mathbf{x}))) - D(\mathbf{x}) + \lambda (\| \nabla_{\mathbf{x}} D(\mathbf{x}) \|_2 - 1)^2$
and 
$L_G = \sum_{\mathbf{x} \in \mathbf{X}} \| \mathbf{x} - G(E(\mathbf{x}))\|_1 - D(G(E(\mathbf{x})))$, respectively.
Lagrange multiplier $\lambda$ is a user specific parameter.
The encoder and generator are trained jointly.

The encoder $E$ is to 
map an input image to its representation in a latent space.
The generator $G$ is to
reconstruct an image back from a latent representation.
Then, to synthesize a counterfactual image $\hat{\mathbf{x}}$, 
a base image $\mathbf{x}$ is encoded to a latent base $\mathbf{z} = E(\mathbf{x})$.
A counterfactual representation $\mathbf{z}^\ast$ is obtained through 
optimization:
$\min_{\hat{\mathbf{z}}} \| \hat{\mathbf{z}} - \mathbf{z}\|_2^2 + \log (1 + \sum_{i=1}^K \exp C_K(G(\hat{\mathbf{z}}))_i )$,
where $C_K(\cdot)_i$ gives the $i^{th}$ output of a $K$-class classifier.
At last, a counterfactual image $\hat{\mathbf{x}}$
is generated: $\hat{\mathbf{x}} = G(\mathbf{z}^\ast)$.
Thus a counterfactual $\hat{\mathbf{x}}$ is presumably foreign, but very similar to its domestic base $\mathbf{x}$.

%

The second term of the counterfactual objective
is supposed to constrain a latent $\hat{\mathbf{z}}$ to be foreign.
Neal et al
have formulated it based on
the assumption that 
a classifier $C_K$ gives low output values on a foreign sample.
This assumption glimpses that LC and counterfactual approaches
could complement each other and the issue is worth a dedicated study. 

With a concern that some information might have been lost
during supervised training,
Yoshihashi et al\cite{Yoshihashi_2019_CVPR} 
have proposed classification-reconstruction learning, where a compact representation is learned simultaneously to the classification.
The compact representation  $\mathbf{z}$ is learned through reconstruction of the input.
Then, it is used as additional information (along with classifier's penultimate vector $\mathbf{a}$) provided to a foreign detector---a binary classifier based on distance between $(\mathbf{z}, \mathbf{a})$ and their average values.
They believe that the compact representation will compensate for the presumably lost information.

In spite of these approaches,
OSR remains greatly challenging \cite{BoultEtAl2019}.
Most OSR approaches require a considerable 
extra mechanism
or a re-design of the entire model.
%
In the sense of a substantial effort crafted for the task,
they are more comparable to Kahneman\cite{kahneman2011thinking}'s analytical system II decision.
LC approach requires much less effort
and
relies on quick deduction using only cues provided by a base classifier.
It is more analogous to Kahneman's instinct system I decision.
As human survival relies on both decision systems,
we believe that a practical OSR system, or more generally a robust intelligent agent, may not need to pick only one best approach.
Both systems can co-exist and complement each other, as this has shown
to be a winning strategy in nature.
A more comprehensive review on OSR is provided by Geng et al\cite{GengEtAl2020}.

\paragraph{Concern over information loss}
A concern over information loss\cite{Yoshihashi_2019_CVPR} might be associated to softmax bottleneck\cite{yang2018breaking, KanaiNEURIPS2018}.
Yang et al\cite{yang2018breaking} 
have analysed 
a softmax-based model for its capacity to represent a conditional probability for a language model.
Based on matrix factorization, they have concluded that a softmax-based model does
not have enough capacity to express the true language distribution.
This is referred to as ``softmax bottleneck''.

Their rationale is drawn based on the diversity of a domain of natural language
and a standard practice of computing a softmax language model.
A softmax language model is computed
using 
a logit, which is
a dot product between a fixed-size context vector and a word-embedding vector.
The chosen fixed size is generally too small
for diversity of a natural language context.
How this nature carries over to other domains and settings
may be subject to dedicated studies.
Nonetheless, our investigation on LC effectiveness may answer this concern for OSR in some degree.

Related but with slightly different objectives,
many studies\cite{HainesXiang2014,GalGhahramani2016,OberdiekEtAl2018}
investigate mechanisms to quantify
uncertainty of a classification inference.
Inference uncertainty is a measure
quantifying a degree of confidence or a level of expertise
in making a particular prediction.
Gal and Ghahramani\cite{GalGhahramani2016} have discussed that inference uncertainty is different from a model confidence, which is conventionally taken as a value of each softmax output.
A value of the softmax output can be shown to be very high (close to one) even when the input lies far beyond vicinity of the training samples in the input space.
In this respect, quantifying inference uncertainty is similar to quantifying a degree of being foreign 
(in our context).
However, a striking distinction between identifying a foreign
and quantifying uncertainty
is 
at the difficult classification or ambiguity among domestic classes.
Difficulty in distinguishing among domestic classes 
is well encompassed by inference uncertainty, 
but this is not an issue of foreign identification.
However, as these are closely related, a potential application of LC to inference uncertainty seems highly likely
and our study here could lay a ground for such an investigation.

Working on active learning criteria to select unlabeled samples from a data pool to ask experts,
Haines and Xiang\cite{HainesXiang2014} propose a soft selection strategy
along with 
approximation using Dirichlet process.
The strategy is that
samples 
with high approximate misclassification probabilities 
are more likely to be selected.
Haines and Xiang
choose
a misclassification probability
over an entropy---commonly used in active learning---
for that
an inclusion of a new class
poses a great challenge for 
an efficient application of an entropy.
%
Given an input $\mathbf{x}$, 
approximate
misclassification probability 
$p(\mathrm{wrong}|\mathbf{x}) = 1 - P_n(k'|\mathbf{x})$,
where
$P_n(k'|\mathbf{x}) \equiv p(y = k'|\mathbf{x})$;
$k' = \arg\max_{k \in C} P_c (k|\mathbf{x})$
when $C$ is a set of all domestic classes
and $P_c (k|\mathbf{x})$ is a probability calculated by a classifier, i.e., a softmax output.
Based on Dirichlet process---specifically a chinese restaurant process---, 
$P_n(k|\mathbf{x})$ can be obtained through normalizing
Haines and Xiang deduction: 
$P_n(k|\mathbf{x}) \propto \frac{M_k}{\alpha + \sum_{i \in C} M_i} P_c(\mathbf{x}|k)$ if $k \in C$
and  $P_n(k|\mathbf{x}) \propto \frac{\alpha}{\alpha + \sum_{i \in C} M_i} P(\mathbf{x})$ if $k$ is a new class,
where
$M_k$ is a number of instances labelled with class $k$.
Haines and Xiang
obtain
$P_c(\mathbf{x}|k) \equiv p(\mathbf{x}|k)$ 
and
$P(\mathbf{x}) \equiv p(\mathbf{x})$  
through 
kernel density estimation.
Parameter $\alpha$ is a concentration coefficient of Dirichlet process.
It can be user-specific,
but 
Haines and Xiang\cite{HainesXiang2014} 
use prior
$\Gamma(1,1)$
and Gibbs sampling to estimate its value.
Haines and Xiang
work on criteria to select data from a data pool.
Therefore, they have all data available for $p(\mathbf{x}|k)$
and $p(\mathbf{x})$.
This is generally a different situation from OSR.
In addition, resorting to density estimation, 
a scalability aspect of this approach remains highly challenging.

\section{Open-Set Recognition via Latent Cognizance}

Since marginalized cognizance is proportional to probability of being domestic \eqref{eq: LC sum g},
Latent Cognizance (LC) can be straightforwardly applied to Open-Set Recognition (OSR).
For OSR with $K$ domestic classes,
our application is as follows.

\begin{enumerate}
\item Choose a base classifier
$C: \mathbf{x} \mapsto \kappa$
where a predicted class $\kappa = \arg\max_{k \in \{1, \ldots, K\}} y_k$
and softmax output $y_k = \mathrm{softmax}_k(\mathbf{a})$
when $\mathbf{a} = [a_1, \ldots, a_K]^T$ is a penultimate vector computed from the input $\mathbf{x}$.
\item Choose a cognizance function $g(a)$.
\item Choose a threshold $\tau$.
\item Compute a marginalized cognizance $c = \sum_{k=1}^K g(a_k)$.
\item If $c \geq \tau$, predict class $\kappa$;
otherwise predict class $0$ (foreign).
\end{enumerate}

Choices of the cognizance function and threshold can be empirically obtained.
A well-adopted classifier can be exploited for the base classifier $C$.
This characteristic is beneficial as this foreign identification can be seamlessly added
to a well-established classification system. 

However, a potential down side is that since LC logic heavily relies on a penultimate vector computed by its base classifier,
its performance may be closely tied to the base classifier.
How much effect the base classifier has on LC had not previously explored.
Thus, we have also investigated this issue in our experiments.

In addition, to properly evaluate OSR,
we propose metric counts (Table~\ref{tbl: TP, FP, and FN}) and a performance metric $Q1$ (based on F-score, \textsection\ref{sec:experiments}).
The metric counts have accounted for every case in OSR evaluation.

\section{Experiments}
\label{sec:experiments}

\begin{figure}[!htb]
	\center{\includegraphics[width=\textwidth] {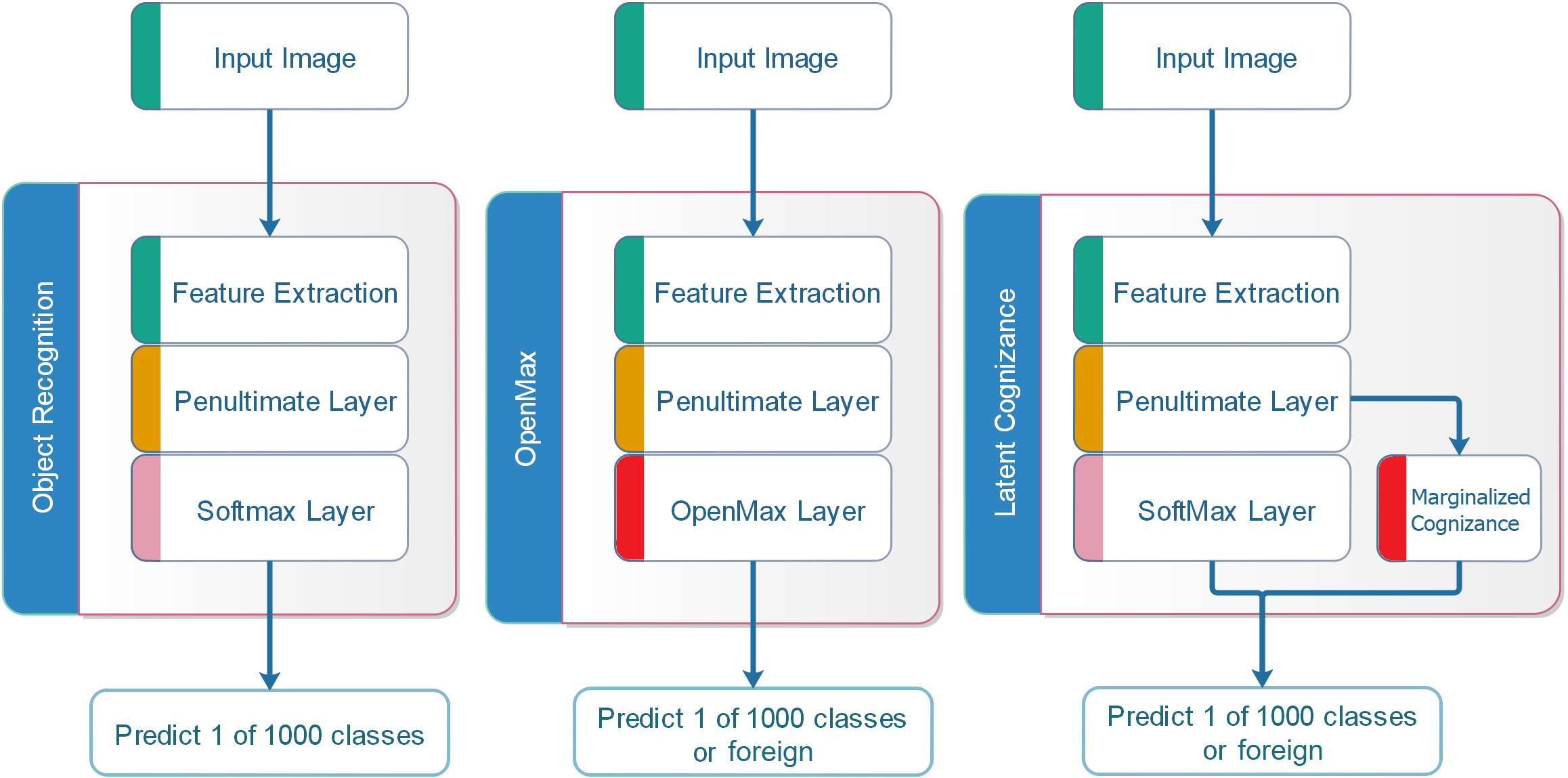}}
	\caption{Architectures of conventional object recognition, OpenMax, and Latent Cognizance.}
	\label{fig:EachApproch}
\end{figure} 

OpenMax and Latent Cognizance (LC) are investigated on Open-Set Recognition (OSR).
%
Fig.~\ref{fig:EachApproch} illustrates structural differences
between OpenMax and LC.
An internal structure of a conventional object recognition network resorts to a softmax layer at the end. 
OpenMax replaces a softmax layer with OpenMax computation. 
LC extends a conventional object recognition with cognizance computation \eqref{eq: LC sum g}.

\paragraph{Model Preparation}
OpenMax and LCs use Alexnet \cite{krizhevsky_imagenet_2017} with pre-trained weights as their base classifier.
A base classifier provides a penultimate vector 
for OpenMax and LC. 
The pre-trained weights were obtained from Caffe 
\cite{jia_caffe_2014}, 
which has trained Alexnet on
the ImageNet Large Scale Visual Recognition Challenge (ILSVRC) 2012 dataset.
There is no additional fine-tuning on Alexnet weights. 


\paragraph{OpenMax hyperparameters}
In its meta-calibration phase (\textsection\ref{sec:OSR}), OpenMax requires domestic learning.
%
Our experiment follows default meta-parameter values of 
Bendale and Boult's implementation\footnotemark
: Weibull tail size $\eta = 20$, a number of top classes $M = 10$, and using Euclidean Cosine method%
.
OpenMax domestic learning is to find $\bm{\mu}_k$, $\tau_k$, $\beta_k$, and $\lambda_k$ for all domestic classes $k = 1, \ldots, K$.
\footnotetext{https://github.com/abhijitbendale/OSDN}
%
Noted that LCs do not require domestic learning. 

\paragraph{LC hyperparameters}
LC uses a cognizance function 
$g(a_k) \sim p(y=k,s|x)$.
The previous work\cite{NakjaiKatanyukul2018a} has empirically explored various candidates for $g(\cdot)$ and found cubic and exponential functions viable. 
Our experiment investigates both as cognizance functions.

\paragraph{Data}
Our experiment uses two datasets to evaluate the models.
%
(1) A domestic test dataset is taken from ILSVRC 2012 validation set, as summarized in Table \ref{table:datsetdescOpenmax_LC}.
It has 50000 images belonging to 1000 classes. 
(2) An open dataset is a combination of 
108360 selected images from ILSVRC 2010
and
15000 fooling images.
All of the 108360 selected examples belong to 360 classes, which all of them are not in ILSVRC 2012.
All of 15000 fooling images are random noise images
with additional
perturbations.
The perturbations are
based on 
an adversarial image generation,  
loosely implemented the work of Szegedy et al\cite{szegedy_intriguing_2013}. 

Szegedy et al's 
formulation is that for a given target class $k$, perform $\min_\mathbf{r}$ $\rVert \mathbf{r} \rVert_2 $ s.t. $f(\mathbf{x}+\mathbf{r})=k$; $\mathbf{x}+\mathbf{r} \in [0,255]^D$, 
where $\mathbf{r}$ is an input perturbation and $\mathbf{x}$ is a base input of $D$ dimensions. 
Function $f(\cdot)$  represents a classifier. 
Our implementation relaxes Szegedy et al's, 
i.e., find $\mathbf{r}: f_k(\mathbf{x}+\mathbf{r}) > \alpha$ 
s.t. $\mathbf{x}+\mathbf{r} \in [0,255]^D$, where $f_k(\cdot)$ is referred to the $k^{th}$ softmax output of the classifier.
Parameter $\alpha$ is user specific and set to $0.9$ in our experiments. 
All generated fooling images have been thoroughly inspected.


\begin{table}[htb]
	\centering
	\caption{Assosicated Datasets.}
	\label{table:datsetdescOpenmax_LC}
\resizebox{\columnwidth}{!}{%
	\begin{tabular}{|l|l|r|l|}
		\hline
Dataset		& Source             & Size & Remark                   \\ \hline
Classifier training set & ILSVRC 2012 training set & 1281167 & Via pre-trained Alexnet \\ \hline
		OpenMax domestic learning        & ILSVRC 2012 training set   & 1281167 & OpenMax only             \\ \hline
		Test domestic data   & ILSVRC 2012 validation set\protect\footnotemark &  50000 &                        \\ \hline
		Test fooling images & Newly generated   & 15000 &  \\ \hline
		Test foreign
		images & ILSVRC 2010 training set   & 108360 & Only classes not in 2012 \\ \hline
		
	\end{tabular}
}
\end{table}

\footnotetext{The ILSVRC 2012 validation set is chosen over the test set for its availability of its ground truth.}

\paragraph{Performance Index}
Our experiment evaluates the models through
performance index $Q1$.
The index is defined as: $Q1=\frac{F_d+F_o}{2}$,
%
where $F_d$ is a performance measure of domestic samples 
and $F_o$ is an F-score of foreign and fooling samples. 

Specifically, $F_d$ is an arithmetic mean of class F-scores,
i.e., $F_d =\frac{1}{K}\sum_{i = 1}^K F_i$, where $F_i$ is an F-score of the $i^{th}$ class and $\{1, \ldots, K\}$ is a set of domestic-class indices.
The class F-scores are defined as 
$F_i = \frac{2 P_i \cdot R_i}{P_i + R_i + \epsilon}$
for $i \in \{1, \ldots, K\}$ and
$F_o = \frac{2 P_o \cdot R_o}{P_o + R_o+ \epsilon}$,
where $\epsilon$ is a small number
for computational stability
and set to 0.0001 in our experiment.
Precisions and Recalls are defined as
$P_i = \frac{TP_i}{TP_i + FP_i+ \epsilon}$
and
$R_i = \frac{TP_i}{TP_i + FN_i}$
for $i \in \{1, \ldots, K\}$ 
and
$P_o = \frac{TP_u + TP_f}{TP_u + TP_f + FP_u+ \epsilon}$
and
$R_o = \frac{TP_u + TP_f}{TP_u + TP_f + FN_u + FN_f}$.
True positives $TP$'s, false positives $FP$'s, and false negatives $FN$'s are defined as shown in Table~\ref{tbl: TP, FP, and FN}.

\begin{table}[htbp]
	\caption{
Our OSR metric.
Symbols $i$, $i'$, $f$, and $u$ represent respectively a sample of domestic class $i$, a sample of domestic class $i'(\neq i)$, a fooling sample, and a sample of any foreign class.
The evaluated systems do not have $f$ output.
Thus, predicting $u$ on $f$ is counted as $TP_f$.}
	\begin{center}
		\begin{tabular}{|l|l|l|}
			\hline
			Ground Truth & Prediction & Metric Count  \\
			\hline
			$i$ & $i$ & $TP_i$ \\
			\hline
			$i$ & $i'$ & $FN_i$ and $FP_{i'}$ \\
			\hline
			$i$ & $u$ & $FN_i$ and $FP_u$ \\
			\hline
			$u$ & $i$ & $FN_u$ and $FP_i$ \\
			\hline
			$u$ & $u$ & $TP_u$ \\
			\hline
			$f$ & $i$ & $FN_f$ and $FP_i$ \\
			\hline
			$f$ & $u$ & $TP_f$ \\
			\hline
		\end{tabular}
		\label{tbl: TP, FP, and FN}
	\end{center}
\end{table}

\paragraph{Comfort Ratio}
A proportion of domestic data
can be an indicator of
how difficult the task is.
A ratio of domestic samples
to
all samples 
will be called a comfort ratio. 
Our investigation experiments 4 scenarios of 
different comfort ratios, as shown in Table~\ref{tbl: data set up}.
A different number of images per foreign class is chosen 
to set a scenario.

Subsections~\ref{sec:results},~\ref{sec:error analysis}
and~\ref{sec:adversarial}
provide the main results,
error analysis
and additional investigation on potential application to adversarial-image detection.

\begin{table}[htb]
	\caption{Test scenarios.}
	\begin{center}
		\begin{tabular}{|l|r|r|r|r|}
			\hline
			Test Case& I & II & III & IV \\
			\hline
			Comfort ratio    & 0.625 	& 0.500  & 0.333  & 0.288 \\ 
			\#images/foreign class     & 42       & 97     & 236    & 300  \\
			Foreign data           & 15120 	& 34920 & 84960 & 108000 \\
			Fooling data          & 15000   & 15000 & 15000 & 15000 \\
			Domestic data            & 50000   & 50000 & 50000 & 50000 \\ 
			\hline
		\end{tabular}
		\label{tbl: data set up}
	\end{center}
\end{table}


\subsection{Results}
\label{sec:results}

\begin{figure}[ht]
	
\begin{center}	
		\includegraphics[width=0.5\textwidth] {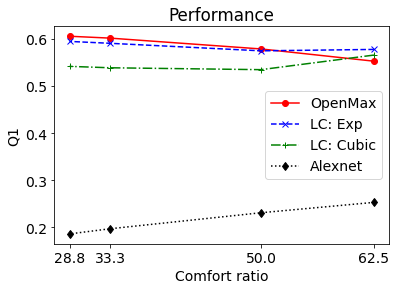}
	\caption{OSR Performance. Q1 of each method
over 
	different comfort ratios. 
OSR performance of Alexnet is provided only for perspective.}
	\label{fig:ResultGraphLCP1andP2}
\end{center}	
	
\end{figure}

Fig.~\ref{fig:ResultGraphLCP1andP2} 
and Tables~\ref{tbl:ResultofOpenmaxP1AndP2Table}
show OSR performance over four scenarios. 
%
%
Table~\ref{tbl:ClassificationTimeTable} reports all time durations spent in each operation. 
Total time spent in domestic learning 
reports a total time spent to fine-tune the open-set capability. 
%
Average time spent in foreign identification is an average time per image that a method spent 
to identify whether an image is domestic or foreign. 

A number in parentheses represents a normalized time. 
It is normalized by classification time per image. 
Classification time per image is an average time Alexnet spent to classify an image. 
It is measured to be $7.52 \times 10^{-2}$ s.
All three methods use Alexnet and are subject to the same classification time per image. 

\begin{table}[ht]
	
	\caption{Performing results}
	\begin{center}
%
\begin{tabular}{|c|l|c|c|c|c|c|}
	\hline
	\multicolumn{2}{|c|}{}                    & 
	\begin{tabular}[c]{@{}c@{}}Comfort \\ Ratio (\%)\end{tabular} & OpenMax & 
	\begin{tabular}[c]{@{}c@{}}Exponential LC \\ $g(a) = \exp(a)$\end{tabular} & 
	\begin{tabular}[c]{@{}c@{}}Cubic LC \\ $g(a) = a^3$\end{tabular} &
	\begin{tabular}[c]{@{}c@{}}Alexnet \\ \end{tabular} \\ \hline
	\multicolumn{2}{|c|}{\multirow{4}{*}{Q1}} & 62.5                                                         & 0.553   & 0.578                                                     & 0.566        & 0.253                                         \\ \cline{3-7} 
	\multicolumn{2}{|c|}{}                    & 50.0                                                           & 0.579   & 0.575                                                     & 0.535        & 0.231                                                \\ \cline{3-7} 
	\multicolumn{2}{|c|}{}                    & 33.3                                                         & 0.602   & 0.591                                                     & 0.539        & 0.197                                               \\ \cline{3-7} 
	\multicolumn{2}{|c|}{}                    & 28.8                                                         & 0.606   & 0.595                                                     & 0.542        & 0.186                                               \\ \hline
\end{tabular}
\end{center}
	\label{tbl:ResultofOpenmaxP1AndP2Table}
\end{table}

\begin{table}[ht]
	\caption{Time spent.
All time durations are reported in seconds.
Normalized time durations
are shown in parentheses.
}
	\begin{center}
		\begin{tabular}{|l|l|l|} 
			\hline
			& Total time spent                                                       & Average time spent \\
			& in domestic learning
			& in foreign identification \\
			\hline
OpenMax  
			& $4.1\times 10^{4}$ ($5.5\times 10^5$) 			
			& $1.03$ ($13.7$)      			
			\\ \hline
Exponential LC   
			& $0$      
			& $7.2\times 10^{-5}$ ($9.6\times 10^{-4}$)   	
			\\ \hline
Cubic LC 
			& $0$      
			& $9.3\times 10^{-5}$ ($1.2\times 10^{-3}$) 		
			\\ \hline
		\end{tabular}
		\label{tbl:ClassificationTimeTable}
	\end{center}
\end{table}

Exponential LC seems to provide slightly better performance than its cubic counterpart.
The performances of all three methods are comparable, 
but LC methods spent considerably less time than OpenMax did. 
In addition, OpenMax requires significant domestic learning time, 
while both cubic and exponential LCs can work right off the shelf. 
All three methods seem to be robust against various comfort ratios. 

\subsection{Error Analysis}
\label{sec:error analysis}

Tables~\ref{tb:OpenmaxMaxQ1result}~and~\ref{tb:LCExponentialMaxQ1result} 
show confusion matrices of OpenMax and exponential LC with thresholds at maximal Q1's. 
The test data is composed of three distinct groups,
while prediction is only limited to either domestic or foreign.
The tables differentiate predicting domestic on domestic samples with 
correct classification (CR) and incorrect classification (IC).
%
Table entries are obtained from:
$CR = \sum_{i=1}^K TP_i$ (predicting a correct class on domestic samples),
$IC = \sum_{i=1}^K FP_i - FN_u - FN_f$ (predicting an incorrect class on domestic samples)
and other metrics
are obtained as specified in Table~\ref{tbl: TP, FP, and FN}.
%

Since OSR performance incorporates both classification and foreign identification aspects, 
Table~\ref{tb:Accruacies seen_unseen} shows separated accuracies by sub-function: 
accuracies of foreign identification (denoted ``F ACC'') and accuracies of classification (denoted ``C ACC''). 

\begin{table}[htb]
	\caption{OpenMax confusion matrices.}
	\label{tb:OpenmaxMaxQ1result}
	
	\centering
	\begin{tabular}{|l|l|l|l|l|}
		\hline
		\multirow{2}{*}{\begin{tabular}[c]{@{}l@{}}Comfort\\ Ratio (\%)\end{tabular}} & \multicolumn{1}{c|}{\multirow{2}{*}{Prediction}} & \multicolumn{3}{c|}{Data}                                                     \\ \cline{3-5} 
		& \multicolumn{1}{c|}{}                            & Domestic                                                       & Fooling & Foreign \\ \hline
		\multirow{2}{*}{62.5\%}                                                      & Domestic                                             & \begin{tabular}[c]{@{}l@{}}CR: 19981\\ IC: 4526\end{tabular} & 1315    & 3957   \\ \cline{2-5} 
		& Foreign                                           & 25493                                                      & 13685   & 11163  \\ \hline
		\multirow{2}{*}{50.0\%}                                                        & Domestic                                             & \begin{tabular}[c]{@{}l@{}}CR: 19755\\ IC: 4322\end{tabular} & 1090    & 8862   \\ \cline{2-5} 
		& Foreign                                           & 25923                                                      & 13910   & 26058  \\ \hline
		\multirow{2}{*}{33.3\%}                                                      & Domestic                                             & \begin{tabular}[c]{@{}l@{}}CR: 18043\\ IC: 3164\end{tabular} & 297     & 16987  \\ \cline{2-5} 
		& Foreign                                           & 28793                                                      & 14703   & 67973  \\ \hline
		\multirow{2}{*}{28.8\%}                                                      & Domestic                                             & \begin{tabular}[c]{@{}l@{}}CR: 17778\\ IC: 3010\end{tabular} & 224     & 20941  \\ \cline{2-5} 
		& Foreign                                           & 29212                                                      & 14776   & 87059  \\ \hline
	\end{tabular}

\end{table}

\begin{table}[htb]
	
	\caption{Exponential LC confusion matrices.}
	\label{tb:LCExponentialMaxQ1result}
	\renewcommand\arraystretch{1.2}
	\centering
	\begin{tabular}{|l|l|l|l|l|}
		\hline
		\multirow{2}{*}{\begin{tabular}[c]{@{}l@{}}Comfort\\ Ratio (\%)\end{tabular}} & \multicolumn{1}{c|}{\multirow{2}{*}{Prediction}} & \multicolumn{3}{c|}{Data}                                                        \\ \cline{3-5} 
		& \multicolumn{1}{c|}{}                            & Domestic                                                          & Fooling & Foreign \\ \hline
		\multirow{2}{*}{62.5\%}                                                      & Domestic                                             & \begin{tabular}[c]{@{}l@{}}CR: 24527\\ IC: 11572\end{tabular} & 646     & 8518   \\ \cline{2-5} 
		& Foreign                                           & 13901                                                         & 14354   & 6602   \\ \hline
		\multirow{2}{*}{50.0\%}                                                        & Domestic                                             & \begin{tabular}[c]{@{}l@{}}CR: 20936\\ IC: 7561\end{tabular}  & 4       & 12226  \\ \cline{2-5} 
		& Foreign                                           & 21503                                                         & 14996   & 22694  \\ \hline
		\multirow{2}{*}{33.3\%}                                                      & Domestic                                             & \begin{tabular}[c]{@{}l@{}}CR: 18365\\ IC: 5576\end{tabular}  & 0       & 20251  \\ \cline{2-5} 
		& Foreign                                           & 26060                                                         & 15000   & 64709  \\ \hline
		\multirow{2}{*}{28.8\%}                                                      & Domestic                                             & \begin{tabular}[c]{@{}l@{}}CR: 17320\\ IC: 4931\end{tabular}  & 0       & 22132  \\ \cline{2-5} 
		& Foreign                                           & 27749                                                         & 15000   & 85868  \\ \hline
	\end{tabular}

\end{table}

\begin{table}[ht]
	\centering
	\caption{Accuracies of foreign identification (F ACC) and accuracies of classification (C ACC).}
	\label{tb:Accruacies seen_unseen}
\begin{tabular}{|c|l|c|c|c|c|}
	\hline
	\multicolumn{2}{|c|}{}                         & \begin{tabular}[c]{@{}c@{}}Comfort \\ Ratio (\%)\end{tabular} & OpenMax & \begin{tabular}[c]{@{}c@{}}Exponential \\ LC\end{tabular} & \begin{tabular}[c]{@{}c@{}}Cubic \\ LC\end{tabular} \\ \hline
	\multicolumn{2}{|c|}{\multirow{4}{*}{F ACC}} & 62.5                                                         & 0.616   & 0.712                                                     & 0.730                                               \\ \cline{3-6} 
	\multicolumn{2}{|c|}{}                         & 50.0                                                           & 0.641   & 0.662                                                     & 0.639                                               \\ \cline{3-6} 
	\multicolumn{2}{|c|}{}                         & 33.3                                                         & 0.693   & 0.612                                                     & 0.643                                               \\ \cline{3-6} 
	\multicolumn{2}{|c|}{}                         & 28.8                                                         & 0.709   & 0.712                                                     & 0.670                                               
\\ \hline
\multicolumn{2}{|c|}{\multirow{4}{*}{C ACC}}   
& 62.5                                                         & 0.815   & 0.679                                                     & 0.629                                               \\ \cline{3-6} 
	\multicolumn{2}{|c|}{}                         & 50                                                           & 0.820   & 0.735                                                     & 0.658                                               \\ \cline{3-6} 
	\multicolumn{2}{|c|}{}                         & 33.3                                                         & 0.851   & 0.767                                                     & 0.696                                               \\ \cline{3-6} 
	\multicolumn{2}{|c|}{}                         & 28.8                                                         & 0.855   & 0.778                                                     & 0.708                                               \\ \hline
\end{tabular}
\end{table}

Breaking down performance into foreign identification and classification reveals that exponential LC performs pretty well on foreign identification (F ACCs are 0.612 to 0.712 across scenarios). 
When considering foreign identification alone,
both LCs are on par with OpenMax.
The classification aspect is mostly attributed to the base classifier. 

Although all methods employ the same Alexnet as their base classifier, classification accuracies are shown to be varied greatly.
The explanation may be 
that foreign identification changes a number of domestic samples to be evaluated for classification performance.
%
%
For example, when difficult domestic samples get incorrectly identified as foreign, 
this hurts F ACC, 
but it helps C ACC: a number of incorrectly-classified samples is decreased. 
In addition to tail statistics and compact abating probability,
OpenMax does thresholding on the maximal class probability.
This mechanism filters out too low class probability 
and may lead to OpenMax tendency toward predicting foreign.
OpenMax thresholding mechanism may have provided a boost on OpenMax classification accuracies, 
as it could bring C ACC to reach 81.5\%, 
conferring to its base classifier Alexnet's reported top-1 accuracy of 57.1\%.

\begin{figure}[htb]
\begin{center}
	\begin{tabular}{cc}
		\includegraphics[width=0.5\textwidth] {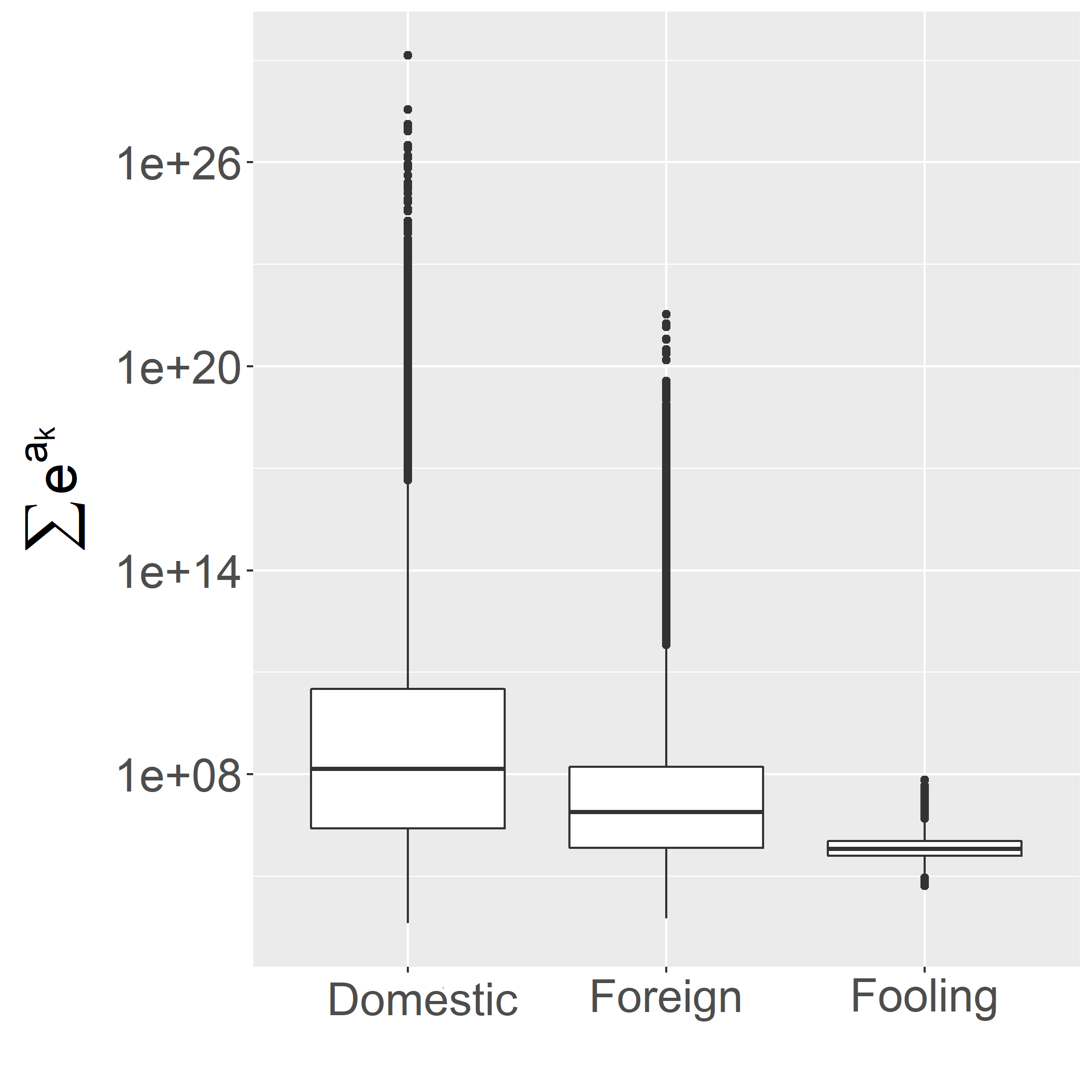}& \includegraphics[width=0.5\textwidth] {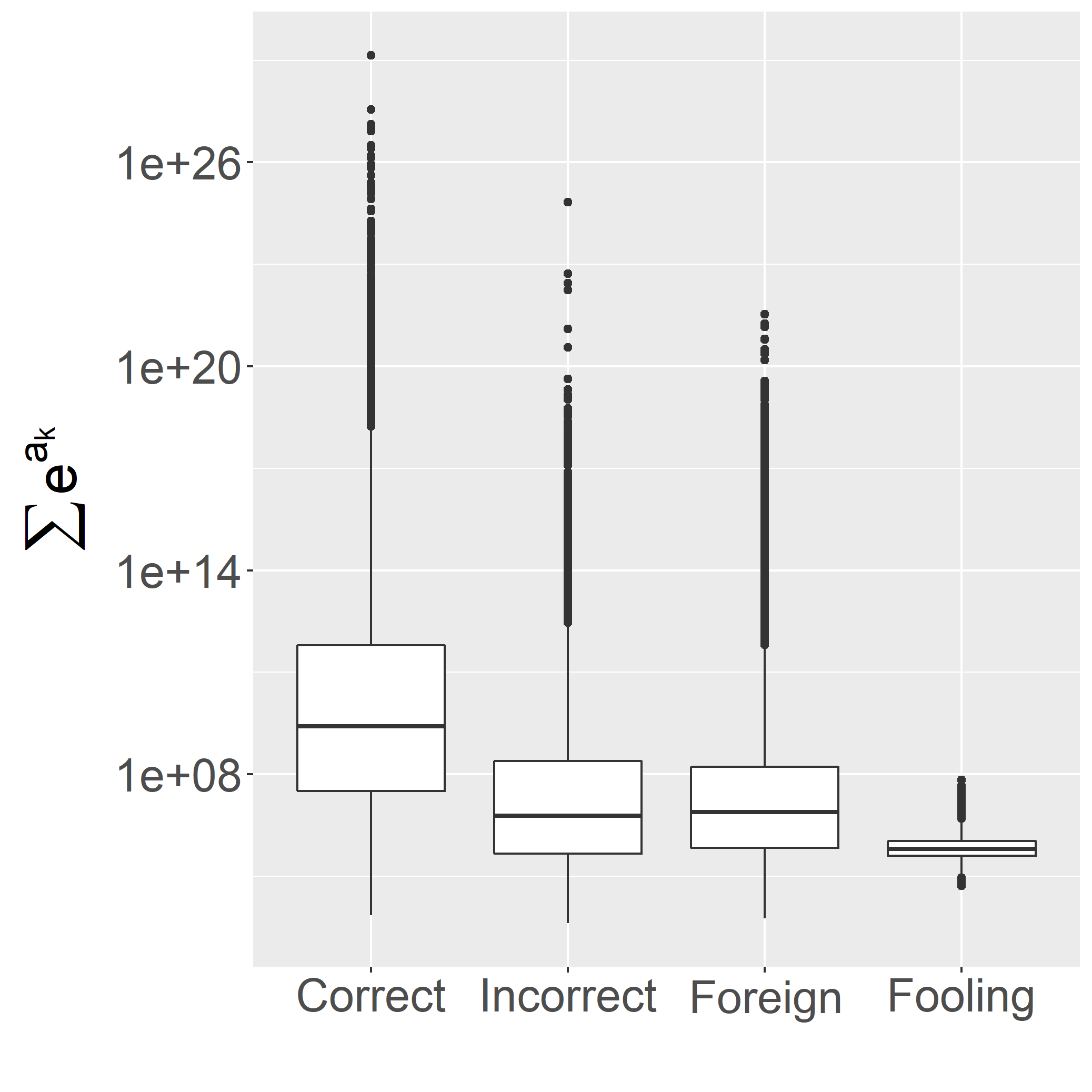} \\
(a) 
&
(b) 
\\
	\end{tabular}

	\caption{Boxplots of marginalized exponential cognizance values of different groups at 28.8\% comfort.
(a) Boxplots are shown for Domestic, Foreign and Fooling groups.	
(b) A domestic group is broken down to correct and incorrect classifications.
	}
	\label{fig:boxplot unseen}
\end{center}
\end{figure}

\paragraph{Exponential LC}
Fig.~\ref{fig:boxplot unseen} shows 
boxplots of marginalized exponential cognizance 
of different data groups at comfort $28.8$\%. 
On the left, marginalized cognizance values of domestic samples (including both correctly-classified and incorrectly-classified domestic samples, denoted ``Domestic''), foreign samples (denoted ``Foreign''), and fooling samples (denoted ``Fooling'') are shown. 
On the right, marginalized cognizance values of correctly-classified domestic samples (denoted ``Correct'') and ones of incorrectly-classified domestic samples (denoted ``Incorrect'') are shown separately.

Fig.~\ref{fig:boxplot unseen} exposes an important aspect for evaluating OSR. 
While it is difficult to threshold for separation between domestic and foreign (as shown in the left plot), 
the use of marginalized cognizance
can well distinguish the correctly-classified domestic samples 
from foreign samples (as shown in the right plot).
A true challenge of OSR 
may actually lie in
differentiation between difficult classifying
and foreign samples,
as a previous work\cite{NakjaiKatanyukul2019a}
has also pointed out.

%
%

\begin{figure}[!htb]	
\begin{center}
		\includegraphics[width=0.5\textwidth] {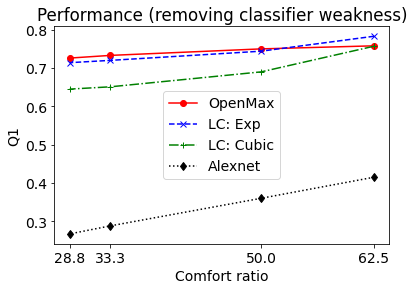}
	\caption{OSR performance over different comfort ratios after removing base-classifier weakness.}
	\label{fig:ResultGraphLCP1andP2OnlyCorrectClassified}
\end{center}
\end{figure}

\paragraph{Evaluating OSR without the Base-Classifier Misclassified} 
To quantify the effect of classifier misclassification,
we examine OSR performance without incorrectly-classified samples.
Table~\ref{tbl:ResultofCorrectPredictionQ1andQ2detail}
and Fig.~\ref{fig:ResultGraphLCP1andP2OnlyCorrectClassified}
show OSR performances of the three methods after removing the base-classifier misclassified samples.
That is, the evaluation was conducted in a similar manner as described earlier, 
but all domestic test samples that Alexnet
misclassified were discarded.
All results seem much more promising:
all Q1 
measures are over 0.6.
%
With improvement over 19\%,
the significance of 
a base classifier is apparent.

%
%
A large number of incorrectly-classified samples
may reflect 
ambiguity in domestic classes
or immaturity of a classifier.
Attention to this aspect may allow
an understanding in the underlying factors
and a further improvement.


\begin{table}[ht]
	
	\caption{Q1 after removing base-classifier misclassified samples. 
Percentage improvement (conferred to Table~\ref{tbl:ResultofOpenmaxP1AndP2Table}) is shown in parentheses.}
\begin{center}
\resizebox{\columnwidth}{!}{%
		\begin{tabular}{|c|lc|cl|cl|cl|cl|}
			\hline
\begin{tabular}[c]{@{}c@{}}Comfort \\ Ratio (\%)\end{tabular} & \multicolumn{2}{|c|}{OpenMax} &
\multicolumn{2}{|c|}{\begin{tabular}[c]{@{}c@{}}Exponential LC \\ $g(a) = \exp(a)$\end{tabular}} &
\multicolumn{2}{|c|}{\begin{tabular}[c]{@{}c@{}}Cubic LC \\ $g(a) = a^3$\end{tabular}} &
\multicolumn{2}{|c|}{\begin{tabular}[c]{@{}c@{}}Alexnet \\ \end{tabular}} \\ \hline

62.5                                                         & 0.758  & (37.1\%)
& 0.783  & (35.5\%)                                                  
& 0.757  & (33.7\%)      
& 0.415  & (64.0\%)                                       
\\ \cline{1-9}
50.0                                                           & 0.750  & (29.5\%)
& 0.744  & (29.4\%)                                                   
& 0.690  & (29.0\%)
& 0.360  & (55.8\%)                                              
\\ \cline{1-9}
33.3                                                         & 0.733  & (21.8\%)
& 0.720  & (21.8\%)                                            
& 0.651  & (20.8\%)
& 0.288  & (46.2\%)                                             
\\ \cline{1-9}
28.8                                                         & 0.726  & (19.8\%) 
& 0.714  & (20.0\%)                                            
& 0.645  & (19.0\%)      
& 0.267  & (43.5\%)                                             
\\ \hline
%
		\end{tabular}
	}
\end{center}
	\label{tbl:ResultofCorrectPredictionQ1andQ2detail}
\end{table}

\subsection{Examining Potential Application to Detection of Adversarial Images}
\label{sec:adversarial}

As exponential LC is shown to accurately identify fooling images
in Tables~\ref{tb:LCExponentialMaxQ1result},
it may appear 
as if LC may be able to address 
the issue of adversarial-image detection.
 
However, 
fooling images are quite different from the actual adversarial images.
The fooling images
were generated using random noise as base images
and this may give away too much clue than actual adversarial images do. 
To properly examine the issue,
15000 adversarial images were generated 
and tested against the 50000 domestic images.
The adversarial images were generated
in the same process generating fooling images
described earlier, 
but---instead of random noise---the base images 
were randomly chosen from images of other 999 classes (excluding the target class).
The resulting images were visually inspected.
 
\begin{figure}[!htb]	
\begin{center}
	\begin{tabular}{cc}
		\includegraphics[width=0.5\textwidth] {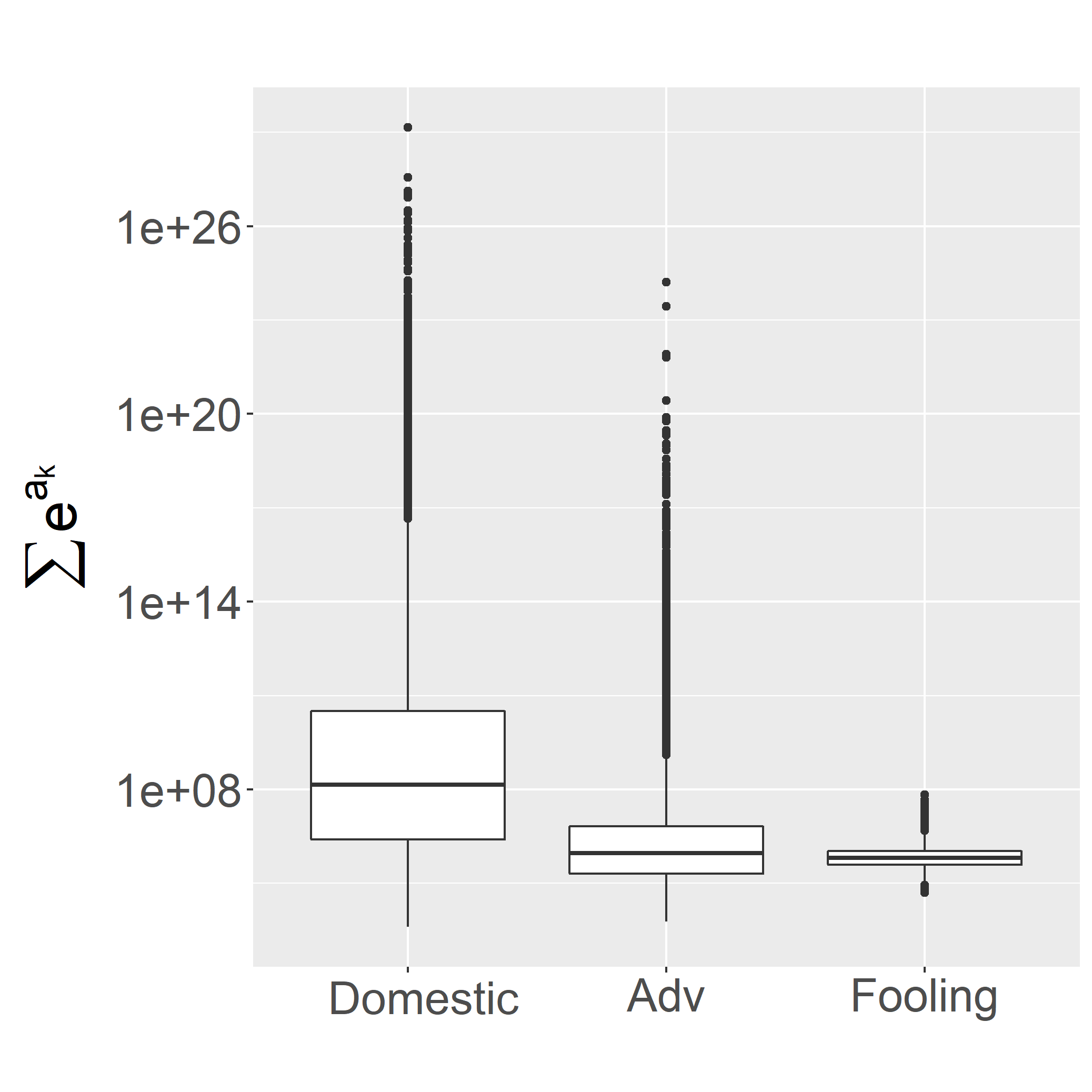} & \includegraphics[width=0.5\textwidth] {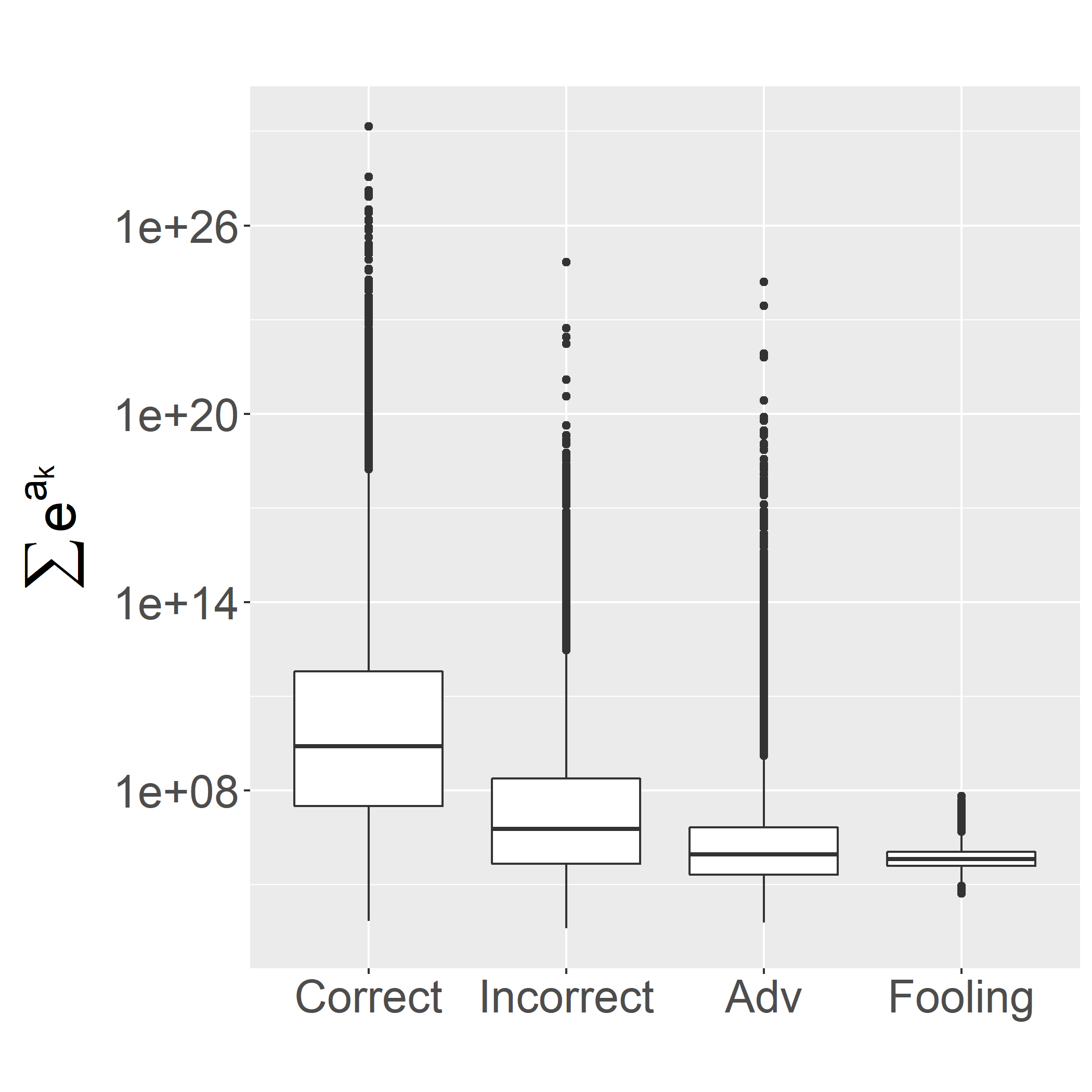} \\
		(a) & (b) \\
	\end{tabular}
	\caption{Boxplots of marginalized cognizance. 
	(a) Boxplots are shown in Domestic, Adv (adversarial), and Fooling groups.
    (b) The Domestic group is broken down to Correct and Incorrect groups.}	
	\label{fig: Adv Boxplot}
\end{center}	
\end{figure}

\begin{figure}[!htb]	
\begin{center}
	\begin{tabular}{cc}
		\includegraphics[width=0.5\textwidth] {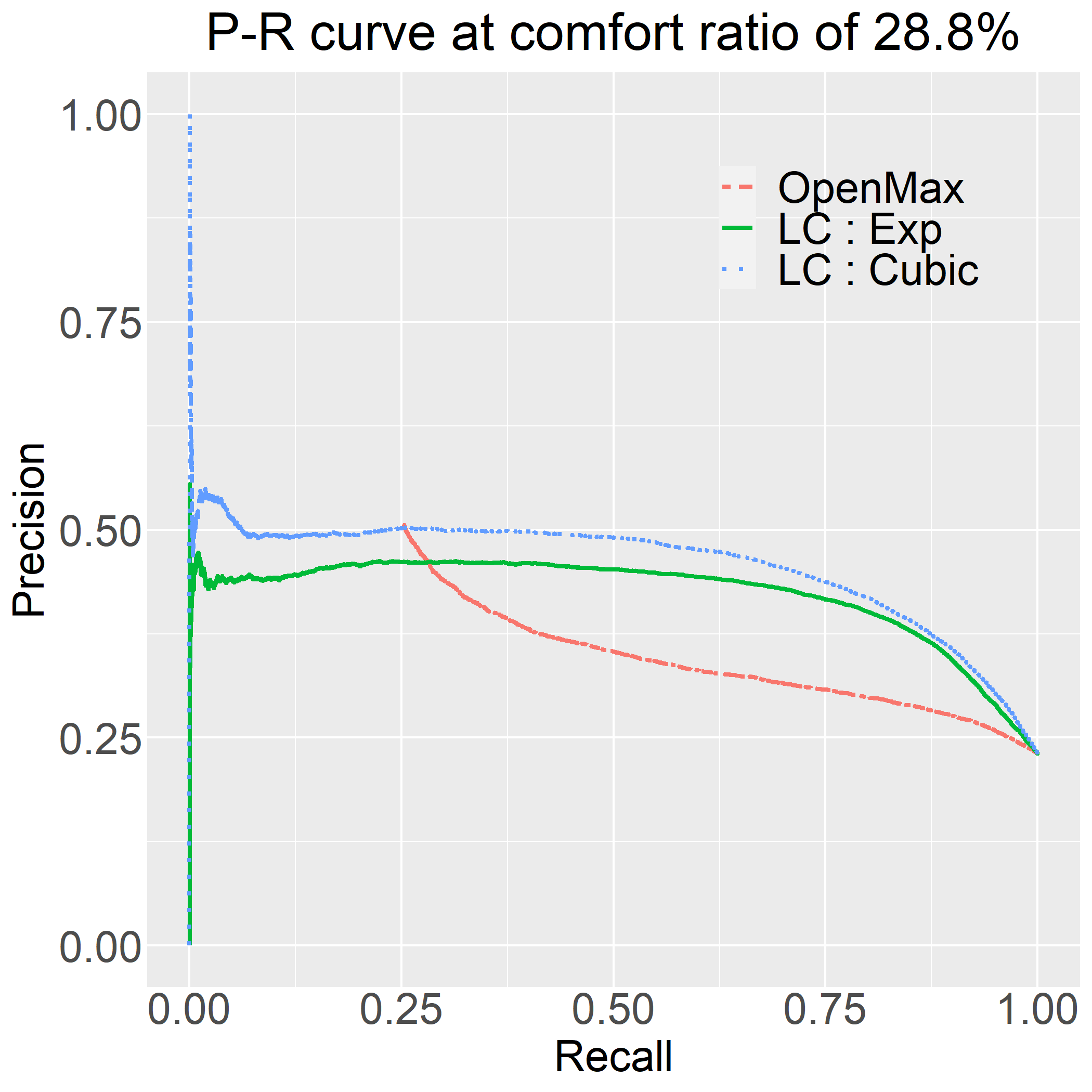} & 
		\includegraphics[width=0.5\textwidth] {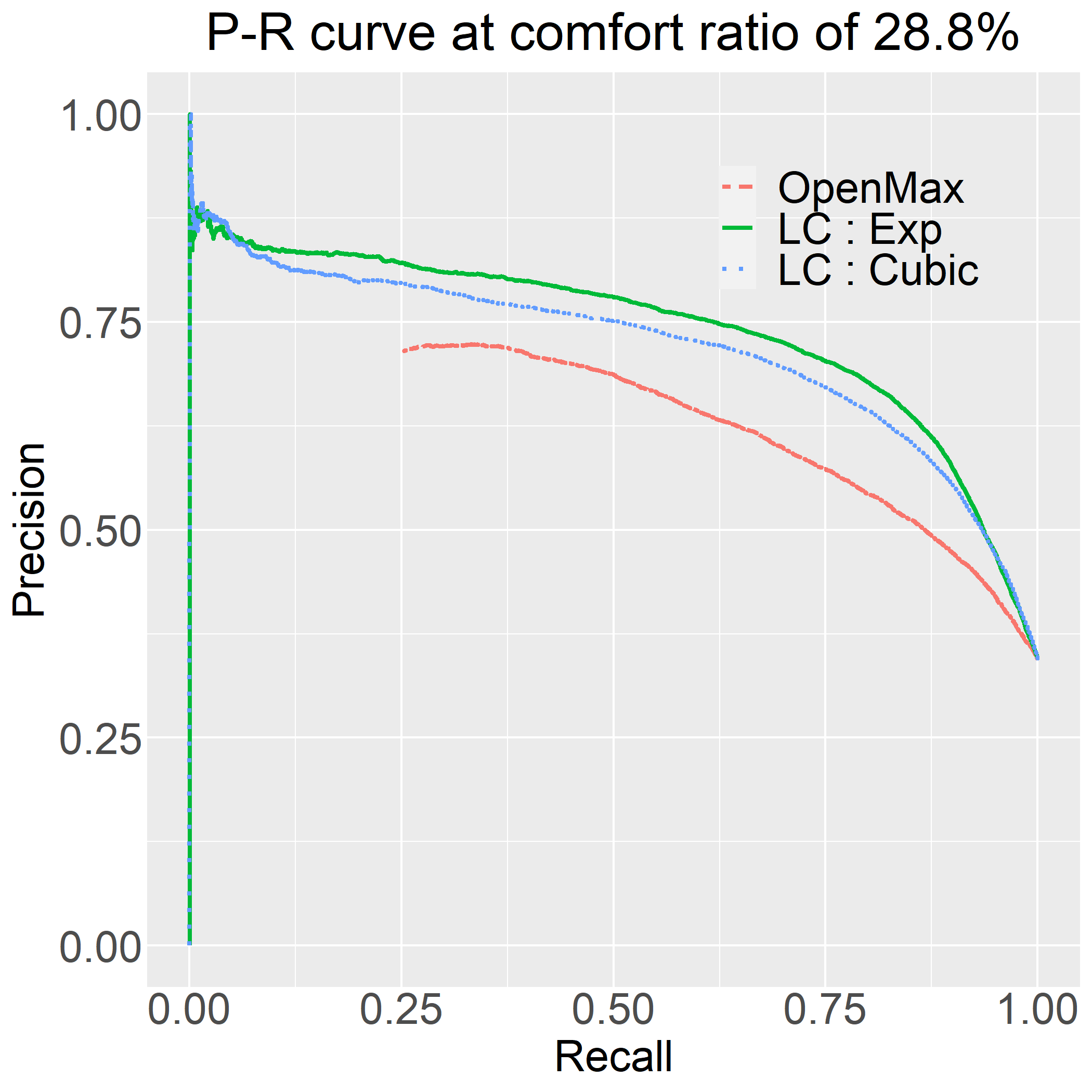} \\
		(a) Domestic & (b) Correct \\
	\end{tabular}
	\caption{Precision-Recall plots of adversarial-image detection.
	(a) P-R curves of distinguishing adversarial images from domestic images.
	(b) P-R curves of distinguishing adversarial images from correctly-classified domestic images.}
	\label{fig: Adv PR}
\end{center}
\end{figure}

Fig.~\ref{fig: Adv Boxplot}
shows boxplots of marginalized exponential cognizance of 
various image types, including adversarial images. 
Fig.~\ref{fig: Adv PR}
shows Precision-Recall (P-R) plots of 
adversarial-image detection: 
binary classification whose positive refers to 
an adversarial sample
and negative refers to a regular example (without adversarial manipulation). 
Table~\ref{tbl: Adversarial AUC} shows 
Area Under Curves (AUCs) of P-R plots of each method.
For perspective, a random classifier was tested on adversarial-image detection 
and achieved AUC 0.317 on average (10 repeats).

Small AUCs (top row, Table~\ref{tbl: Adversarial AUC}) 
rule out a side benefit of any of these OSR methods
as an effective adversarial-image detector.
However, better AUCs (bottom row) are achieved when tested against only correctly-classified samples.
This may 
disclose some potential of these approaches,
but 
an improvement or further investigation may require a dedicated study.

\begin{table}[!htb]
	\caption{AUC of P-R plot: Detecting Adversarial Images.}
	 
	\begin{center}
		\begin{tabular}{|l|c|c|c|c|}
		\hline
		        & OpenMax & Exponential LC & Cubic LC \\
		\hline
		Adversarial and domestic data    & 0.379   & 0.423   & 0.458    \\
		\hline  
		Adversarial and correctly classified data & 0.635   & 0.741   & 0.719    \\
		\hline
		\end{tabular}
		\label{tbl: Adversarial AUC}
	\end{center}
\end{table}




\subsection{Complementary Comparison}

In order to put OSR into perspective, OpenMax and LCs are compared against an off-the-shelf object detection.
An object detection can simply be seen as $F: \mathbf{x} \mapsto \{ \mathbf{b}_1, \ldots, \mathbf{b}_M \}$,
where $\mathbf{x}$ is an input image;
$\mathbf{b}_i$ is the $i^{th}$ detection;
and $M$ is a number of all possible detections.
Each detection $\mathbf{b}_i$ usually composes of bounding box coordinates $\mathbf{c}$, detection score $p$, and object class $k$.
A well-trained 
Faster R-CNN ResNet101 V1 640x640 model%
\footnote{%
\url{https://tfhub.dev/tensorflow/faster_rcnn/ResNet101_v1_640x640/1}, obtained on Oct 7th, 2021.
}%
is chosen for this comparison.

To set up a comparable setting, 
the test dataset composes of 
7020 images.
The 1000 of these images from ILSVRC 2012 validation set belong to 20 classes verified to be domestic for both Alexnet and the object detection.
The 6020 of these images from ILSVRC 2010 belong to 20 classes verified to be foreign for both Alexnet and the object detection.
%
To adapt object detection for OSR, 
the detection score $p$ is treated as a predicted degree of being domestic.

\begin{figure}[!htb]	
\begin{center}
\includegraphics[width=0.5\textwidth] {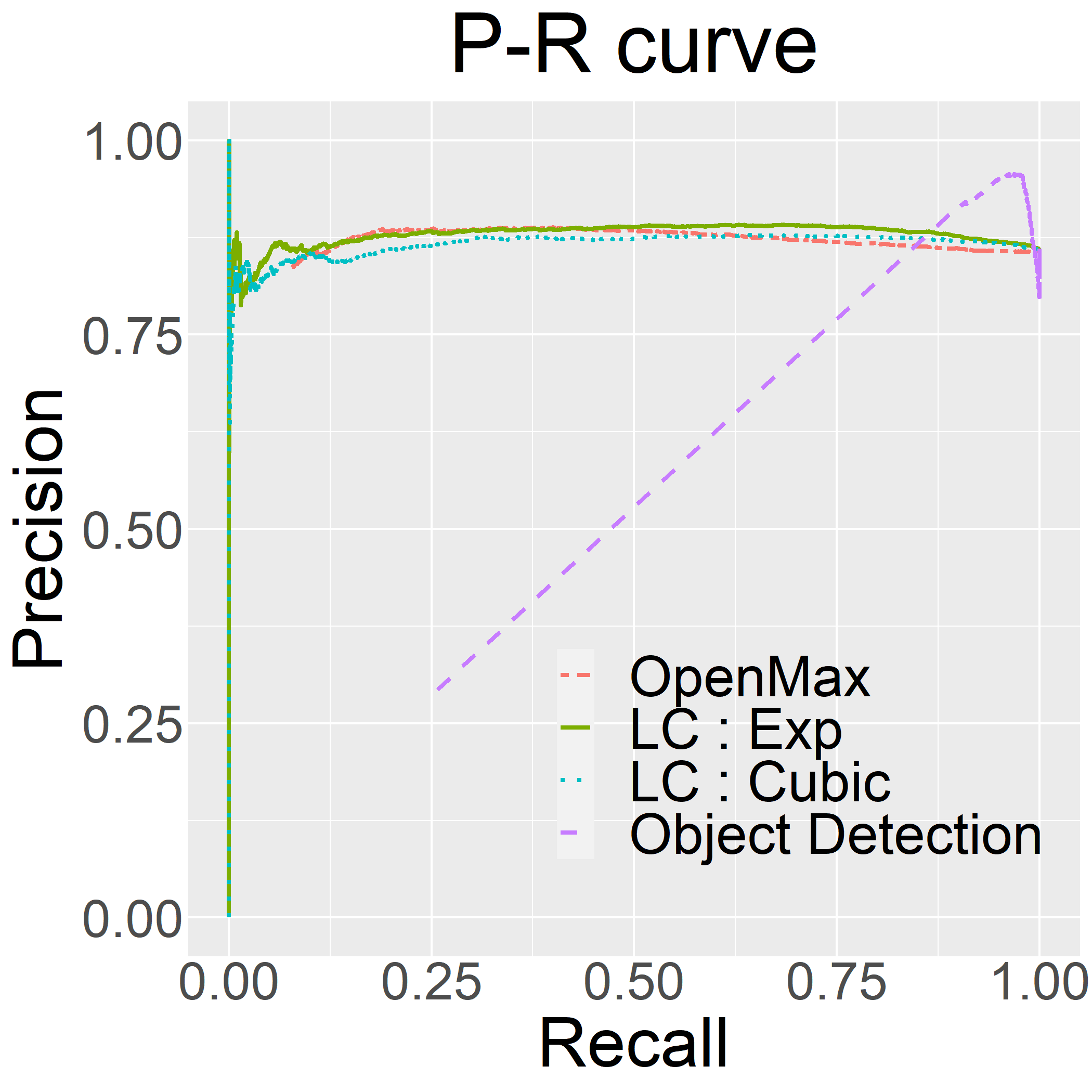} 
	\caption{Precision-Recall plots of foreign detection in the complementary study.}	
	\label{fig: PR comparison OD}
\end{center}	
\end{figure}

The comparison results are provided in Table~\ref{tbl: complementary comparison}.
The P-R plot is shown in Figure~\ref{fig: PR comparison OD}.
The details of this comparison and how the object detection is adapted for OSR are provided in the supplementary materials.
Noted that 
this comparison is only meant to provide a preliminary perspective.
There are factors---such as the underlying models, 
how the models are prepared,
original numbers of classes used in training, 
and how the outputs are interpreted under OSR context%
---that may deserve more attention.
A full potential OSR capability of 
object detection may be worth a dedicated study.

\begin{table}[ht]
	
	\caption{Complementary comparison: dedicated OSR methods and off-the-shelf object detection on OSR (OD-OSR)}
	\begin{center}
\begin{tabular}{|c|c|c|c|c|}
	\hline
Metric & OpenMax & Exponential LC     & Cubic LC        
& OD-OSR \\
       &         & $\sum_i \exp(a_i)$ & $\sum_i a_i^3$	
& $p$  
\\
\hline
Q1     
&	0.6127
& 0.6300
& 0.5767
& 0.6046
\\
\hline
AUC of P-R 
& 0.8713
& 0.8790
& 0.8663
%
& 0.7532
\\
\hline
\end{tabular}
\end{center}
	\label{tbl: complementary comparison}
\end{table}

\section{Conclusion and Discussion}
\label{sec:discussion}

Our investigation has revealed viability of Latent Cognizance (LC) in Open-Set Recognition (OSR)
and re-affirmed the LC underlying hypothesis.
In additions, our study introduces 
performance metric Q1,
discloses some potential on detecting adversarial images,
and shows that a base classifier can affect 19\% or more on the final OSR performance.

\paragraph{OpenMax and LC}
Both LC and OpenMax rely on penultimate values, 
but they use penultimate values differently.
OpenMax uses statistics of penultimate  values
to estimate how far off the ones corresponding to the input are from their statistics.
LC uses only a penultimate vector corresponding to the input
to compute a marginalized cognizance.
LC is simpler to implement, faster to compute,
and yet able to deliver a similar level of effectiveness.

Both LC and OpenMax are quite effective for OSR over a wide range of comfort ratios,
but there are rooms for improvement.
Besides having a better base classifier for both appproaches%
---%
as our study has shown its great influence on the final performance%
---,
OpenMax
is implicitly based on a uni-modal assumption.
Relaxation on this assumption might be a direction to further investigate an OpenMax approach.

For LC,
we see it analogous to Kahneman\cite{kahneman2011thinking}'s System I decision.
Toward OSR, LC could provide a quick judgement on validity of the input at hand.
As human learns to balance decision from both quick/instinctive and slow/analytic systems,
we vision LC or its derivative to work along with a more elaborated approach to provide OSR capability for various scenarios.
It is always better to be prepared for what we might encounter,
but it is nice to have a backup system---the instinctive system---that can work in some degree even for the scenario we might not anticipate.
That is the position at we see LC in the big picture.

\paragraph{Practical issue of LC formulation}
Regarding practical deployment and potential numerical issues,
cubic cognizance may have an issue with negative logit values in the penultimate vector.
Although this was not the case in our experiment, 
but this issue may arise in practice.

Exponential cognizance does not have an issue with negative logits, 
but a large logit value may unstabilize the computation 
(as exponential cognizance could reach numerical infinity for a large logit).
This could be mitigated by a numerically safer application of LC.
For example, $\log( \sum_k \exp(a_k) ) = a_{\max} + \log(\sum_k \exp(a_k - a_{\max}))$ may be a safer version than $\sum_k \exp(a_k)$ used in our experiment.

Note that this log formulation may help explain Nakjai and Katanyukul\cite{NakjaiKatanyukul2018a}'s experiment
that
thresholding on $a_{\max}$ 
gives similar results 
to thresholding on marginalized exponential cognizance.
That is because
logarithm is an increasing monotonic function%
---making thresholding on marginalized exponential cognizance
similar to doing on log marginalized exponential cognizance%
---and the term
$\log(\sum_k \exp(a_k - a_{\max}))$ is likely to be small.


\paragraph{On the concern over information loss}
Effectiveness of LC on OSR may have somewhat assuaged the concern over information loss discussed in Yoshihashi et al\cite{Yoshihashi_2019_CVPR}.
%
%
In addition, marginalized exponential cognizance
has been satisfactorily used
to quantify a degree of being foreign in a counterfactual approach\cite{Neal2018ECCV}, as Neal et al use logit values for their classifier output (see \textsection\ref{sec:background}). 
Our results along with effectiveness shown in previous works\cite{Neal2018ECCV, NakjaiKatanyukul2018a, AtsawaraungsukEtAl2021a}
have supported the hypothesis underlying LC.

\paragraph{OSR metric}
Q1 is based on averaging over performances of classification and foreign detection.
It accounts for all cases (Table~\ref{tbl: TP, FP, and FN}).
The rationales are justified (from our current point of view), 
but an assessment on this and other OSR metrics should be properly studied in their own right.

Additionally, metric Q1 measures the overall OSR performance,
but 
we found that it is more beneficial to examine the break-down performances: classification and foreign identification.
A unified metric that provides a more convenient way to
examine overall OSR performance as well as its underlying factors
could be greatly useful.

\paragraph{Adversarial Detection and Inference Uncertainty}
Although there might be some potential for adversarial detection,
at this point we do not see adversarial detection as a promising side benefit of these OSR approaches.

However, when domestic examples are broken down into correctly and incorrectly classified examples
(Fig.~\ref{fig:ResultGraphLCP1andP2OnlyCorrectClassified}b and~\ref{fig: Adv Boxplot}b), the results seem more favourable
from a perspective of inference uncertainty.
These may suggest that while we are testing LC for foreign and adversarial detection,
it may naturally be more suitable for providing inference uncertainty.
In that case, the outcomes look much more decisively positive.
Re-purposing LC for inference uncertainty
does not lose its value as a mechanism for machine awareness.
A lower value of marginalized cognizance indicates 
that
the classification prediction is likely to be incorrect
for either a wrong class, a foreign, or even an adversarial input.
It still provides an awareness of the input being beyond a machine capability. 

\paragraph{OSR and Object Detection}
Comparing intrinsic OSR methods to object detection modified for OSR
reveals
a marginal benefit of dedicated OSR approaches
over a simply modified object detection.
Although the comparison 
is in a preliminary stage, 
this shows 
strong potential of extending object detection capability to address OSR.
Since the mechanism behind many object detection systems also employs softmax, 
it is possible to enhance this capacity through LC.
A further study on enhancing OSR capacity of object detection
could benefit both domains.

However, object detection has been trained on a different setting.
Objects of a ``foreign'' class might accidentally be in some of the training images, but their class is not one of the target classes.
(This situation makes it difficult to justify a class as foreign
without inspecting all the images.)
Therefore, the model might have learned the presumed foreign objects. 
Consequently, the class is not truely foreign for the model.
Thus, the experiments and evaluation should carefully be conducted.

\paragraph{Big Picture}
Regardless of how OSR is carried out,
it is a crucial step in machine intelligence.
Enhancing classification with foreign identification
is, to a large extent, 
analogous to enriching machine intelligence
with an awareness of its own limitation.
It is the awareness that the question is beyond what a system could answer.
This awareness when fully developed could allow a safer measure against a dynamic and diverse setting on which an intelligent system will be deployed.
Therefore, OSR and similar concepts in other domains 
should be sufficiently addressed for 
the development of a robust intelligent agent. 


%
%
%
%
\bibliographystyle{elsarticle-num}
\bibliography{bibLC,ExportedItems}

\end{document}